\documentclass{article}

\usepackage{arxiv}

\usepackage[utf8]{inputenc} 
\usepackage[T1]{fontenc}    
\usepackage[numbers,compress]{natbib}
\usepackage{hyperref}       
\usepackage{url}            
\usepackage{booktabs}       
\usepackage{amsfonts}       
\usepackage{nicefrac}       
\usepackage{microtype}      
\usepackage[table]{xcolor}  
\usepackage{multirow}
\usepackage{makecell}
\usepackage{graphicx}
\usepackage{amsmath}
\usepackage{amssymb}
\usepackage{array}
\usepackage{listings}

\definecolor{headergray}{RGB}{230,230,230}
\definecolor{ablblue}{RGB}{233,242,251}
\definecolor{ablred}{RGB}{251,237,237}

\title{INTACT: Ego-Guided Typed Sparse Evidence Retrieval for Heterogeneous Collaborative Perception}

\author{%
  Chen Li\textsuperscript{1},
  Shengrong Yuan\textsuperscript{1},
  Jialong Zuo\textsuperscript{1},
  Xinzhong Zhu\textsuperscript{2},
  Nong Sang\textsuperscript{1},
  Changxin Gao\textsuperscript{1}\thanks{Corresponding Author.}
  \\[2pt]
  \textsuperscript{1} National Key Laboratory of Multispectral Information Intelligent Processing Technology,\\ School of Artificial Intelligence and Automation, Huazhong University of Science and Technology,\\
  \textsuperscript{2} Zhejiang Normal University.\\[2pt]
  \texttt{\{lichenrui27, shengrongyuan, jlongzuo, cgao\}@hust.edu.cn}
}

\begin{document}
\maketitle
\begin{abstract}
Collaborative perception extends the perceptual range of autonomous vehicles by sharing information across agents, but heterogeneous sensors and perception models make intermediate feature fusion difficult to deploy at scale. Existing heterogeneous collaboration methods typically follow a translation-first paradigm: collaborator features must be aligned, adapted, or projected into an ego-compatible space before fusion. Such feature-compatibility contracts improve fixed-system performance, but they couple deployment to collaborator-specific adaptation and make newly joined heterogeneous agents costly to integrate. To address this gap, we propose INTACT, an ego-guided typed sparse evidence retrieval framework for heterogeneous collaborative perception. Instead of translating an entire collaborator feature map, INTACT lets the ego vehicle issue typed evidence queries that express suspected objects and evidence-deficient regions. Collaborators respond only with local evidence at queried locations, and the ego selects useful responses through sparse per-query routing and injects them through gated residual write-back. This changes the compatibility requirement from global feature-map interpretability to local, typed response comparability under ego-issued queries, enabling a zero-training heterogeneous insertion protocol in which the ego interface is trained once and new collaborators join through checkpoint merging. Extensive experiments on simulated and real-world heterogeneous collaborative perception benchmarks validate the effectiveness and deployability of INTACT. On OPV2V-H, INTACT achieves 80.1 AP70 with only 0.52M additional parameters and 18.0~$\log_2$ communication volume, corresponding to about 16$\times$ compression over dense feature transmission. On DAIR-V2X, INTACT achieves 43.8 AP50 under challenging real-world conditions.
\end{abstract}

\section{Introduction}

Collaborative perception enables autonomous vehicles to share perceptual information, compensating for occlusions, limited fields of view, and long-range sensing challenges faced by single-agent systems~\cite{arnold2020cooperative,nair2024collaborative}. Intermediate feature fusion has become the dominant paradigm: compared with early fusion and late fusion, it offers a favorable trade-off between perceptual richness and communication cost~\cite{hu2022where2comm,v2x_vit,wang2020v2vnet}. However, this paradigm assumes that the exchanged feature representations are mutually interpretable. In practice, vehicles may carry different sensor suites, such as LiDAR, cameras, or inputs at different resolutions; use different backbone architectures, such as PointPillars, SECOND, EfficientNet, or ResNet; and be trained under different objectives~\cite{lang2019pointpillars,yan2018second,tan2019efficientnet,resnet}. Such heterogeneity produces incompatible feature representations, making cross-vehicle fusion fundamentally difficult~\cite{mpda,heal,stamp}.

Most existing heterogeneous collaborative perception methods share a common premise: heterogeneous representations must first be made compatible before collaboration can occur. We refer to this as the \textit{translation-first} paradigm. Feature alignment methods, such as MPDA~\cite{mpda} and HEAL/BackAlign~\cite{heal}, learn cross-representation mappings between specific ego-collaborator pairs. Shared protocol methods, such as STAMP~\cite{stamp} and GT-Space~\cite{gtspace}, predefine a common representation space into which all agents project their features. Although these methods differ in mechanism, they share a fundamental limitation: the collaboration interface is coupled with feature-level compatibility. Introducing a new collaborator type, with a different sensor, backbone, or training procedure, usually requires learning a new alignment function for alignment-based methods, or ensuring that the predefined protocol space can sufficiently preserve the distinctive information of the new modality for protocol-based methods. This coupling limits deployment scalability.

We argue that this coupling is not necessary. Consider a simple scenario: one participant holds a precise computer-generated drawing, while another holds a hand-drawn sketch of the same scene. The two representations cannot be directly merged, but collaboration does not require converting the entire sketch into the other's drawing format. If the request is specific enough, for example asking whether a structure exists at a location or whether a boundary is reliable, the responder can provide local evidence from its own representation. For heterogeneous collaborative perception, this means that a collaborator does not have to translate its entire feature map into an ego-compatible representation. It only needs to return relevant local evidence according to the query issued by the ego vehicle. The key insight is that the query itself can serve as the collaboration interface: it defines interaction through location, query type, and task context, while abstracting away the internal representation differences among collaborators.
\begin{figure*}[t]
\centering
\includegraphics[width=0.98\textwidth]{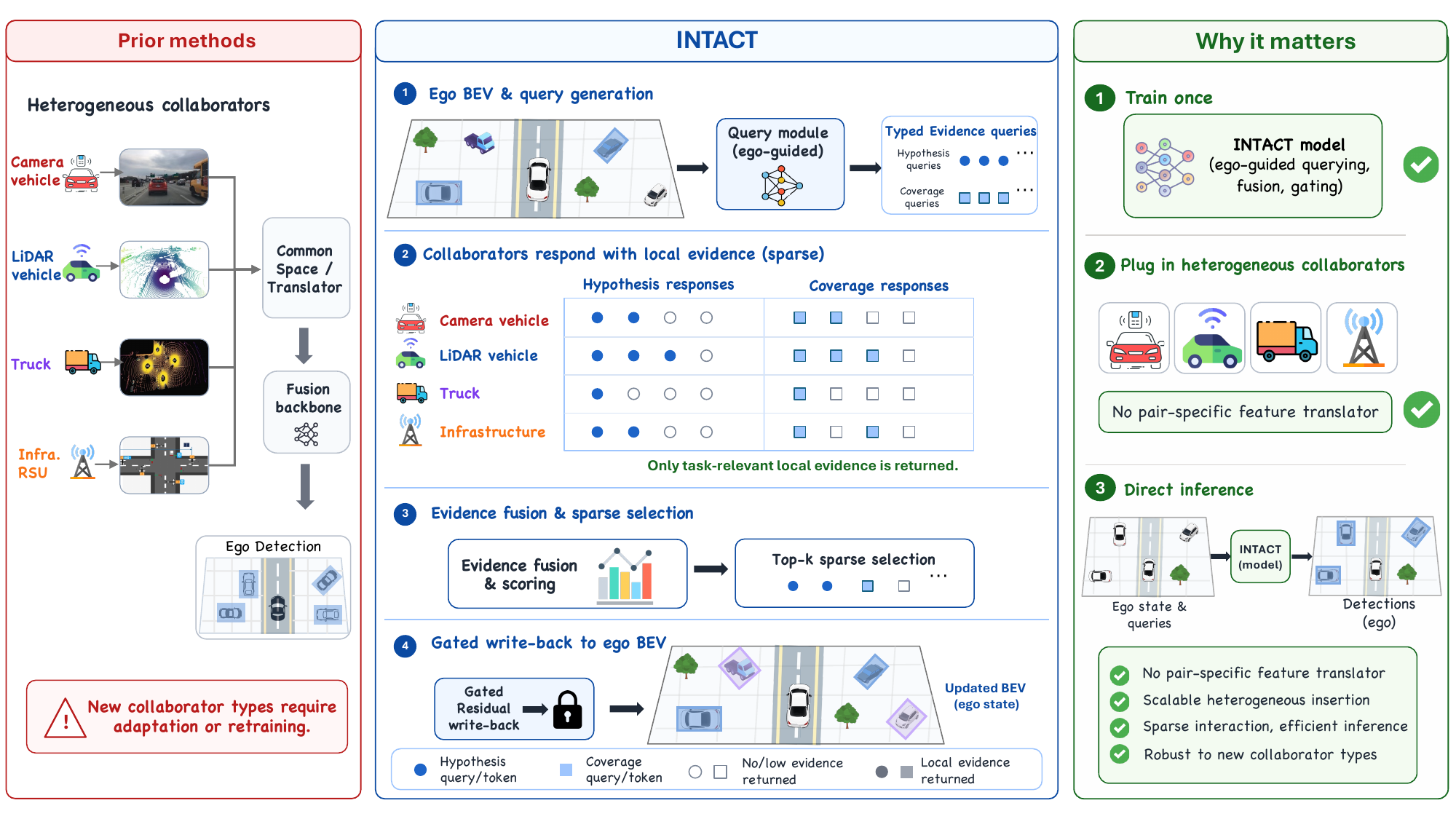}
\caption{Conceptual comparison of heterogeneous collaborative perception interfaces. Prior translation-first methods align, adapt, or project heterogeneous collaborator features into an ego-compatible space before fusion. INTACT instead uses an ego-issued query interface: the ego generates typed evidence queries, collaborators return query-conditioned local responses at query anchors, and selected evidence is injected through gated residual write-back. This avoids pair-specific feature translators for newly inserted heterogeneous collaborators.}
\label{fig:explanation}
\end{figure*}

This observation also aligns with recent query-based or instance-level collaborative perception studies. Prior work shows that, compared with directly exchanging dense features, concentrating collaboration on a smaller number of task-relevant locations or object instances can reduce communication redundancy and improve interaction efficiency~\cite{xu2025instinct}. However, in heterogeneous collaboration, queries should not be viewed only as a sparsification device. As summarized in Fig.~\ref{fig:explanation}, the deeper distinction lies in the definition of the collaboration interface. Prior methods first make heterogeneous collaborator features compatible through translation, adaptation, or projection, and then perform fusion. INTACT instead lets the ego vehicle define the interaction by issuing typed evidence queries from its current BEV state. These queries specify which object hypotheses require verification and which regions lack sufficient evidence, so collaborators only need to return local responses around the queried locations. This changes the role of the ego representation: it is no longer a passive target waiting for external feature fusion, but an active requester that retrieves task-relevant local evidence from heterogeneous collaborators. This shift provides the basis for the train-once and reusable deployment pattern illustrated in Fig.~\ref{fig:explanation}.

To this end, we propose INTACT, an ego-guided typed sparse evidence retrieval framework for heterogeneous collaborative perception without learning pairwise feature translators. The framework organizes cross-vehicle interaction as a query, response, and gated residual write-back loop. The ego vehicle first generates typed evidence queries from its BEV representation to express object hypotheses and evidence-deficient regions. Collaborators then return local feature responses only at queried locations. The ego further selects more relevant evidence from candidate responses and injects it into its own representation through gated residual write-back. Therefore, the collaboration interface is no longer defined by a global feature space, but by local task requests issued by the ego vehicle. Since the interface is defined by ego queries, the model learns how to issue evidence requests, select local responses, and safely inject evidence, instead of learning a separate feature translator for every ego-collaborator combination. As a result, the ego-side interface only needs to be trained once and can then be reused when different heterogeneous collaborators are inserted, avoiding additional pairwise heterogeneous adaptation training.

We conduct extensive experiments on simulated and real-world heterogeneous collaborative perception benchmarks, including OPV2V-H~\cite{heal}, DAIR-V2X~\cite{dair_v2x}, and V2X-Real~\cite{v2xreal}. The results show that INTACT maintains stable detection performance across different heterogeneous configurations while using low parameter and communication overhead. On OPV2V-H, INTACT achieves 80.1 AP70 with only 0.52M additional parameters and 18.0~$\log_2$ communication volume, corresponding to about 16$\times$ compression over dense feature transmission. On DAIR-V2X, INTACT achieves 43.8 AP50 under challenging real-world conditions. These results validate the effectiveness of ego-query-driven local evidence retrieval in terms of accuracy, efficiency, and deployment scalability.

Our contributions are summarized as follows:
\begin{itemize}
    \item We revisit heterogeneous collaborative perception from the perspective of collaboration interface design. Instead of treating feature compatibility as the prerequisite for collaboration, we define the collaboration interface over ego-generated typed evidence queries, shifting heterogeneous collaboration from full-map representation translation to ego-demand-driven local evidence retrieval.

    \item We propose INTACT, an ego-guided typed sparse evidence retrieval framework. It organizes the ego vehicle's current BEV representation, object responses, and evidence-deficient regions into answerable queries, and performs heterogeneous collaboration through a query, response, and gated residual write-back loop: the ego issues queries, collaborators return local evidence, and the system selects useful responses and injects them into the ego representation through gated residual write-back.

    \item We introduce a train-once, reusable ego-side mechanism for heterogeneous collaboration. Different from learning a separate feature translator for each ego-collaborator combination, INTACT learns an ego-defined evidence interaction interface, enabling different heterogeneous collaborators to provide complementary local evidence without additional pairwise adaptation training.

    \item We conduct extensive experiments on simulated and real-world datasets. Results on OPV2V-H, DAIR-V2X, and V2X-Real show that INTACT achieves a favorable balance among detection accuracy, parameter efficiency, communication cost, and deployment scalability, and further analyses verify the effectiveness of its core designs.
\end{itemize}

\section{Related Work}

\subsection{Feature Alignment and Adaptation}
Collaborative perception has traditionally made feature compatibility the premise of intermediate fusion. Homogeneous systems assume shared architectures or feature distributions so that spatially aligned BEV features can be fused after coordinate transformation. Heterogeneous systems relax the sensor and model assumptions but often keep the same underlying contract: before fusion, a collaborator's representation must be made interpretable to the ego. HEAL/BackAlign~\cite{heal} and MPDA~\cite{mpda} are representative of this direction, learning alignment mechanisms or domain adaptation to reduce the representational gap between heterogeneous agents. These designs make heterogeneity explicit, but they still require a learned compatibility mechanism between feature spaces. INTACT avoids this by not asking the collaborator to export an ego-compatible dense feature map; it only requests local responses to typed ego-issued spatial queries and controls their effect through gated residual write-back.

\subsection{Shared Protocols and Generative Communication}
A second line of work introduces structured communication protocols to reduce dependence on raw feature compatibility. STAMP~\cite{stamp} and GT-Space~\cite{gtspace} construct shared representation spaces into which agents project features before interaction. GenComm~\cite{gencomm} and CodeFilling~\cite{cofilling} optimize communication through generative compression or code-based reconstruction. Where2comm~\cite{hu2022where2comm} and Who2com~\cite{who2comm} introduce spatial or agent-level sparsity into collaborative communication. EIMC~\cite{eimc} uses instance-aware top-K messaging for multi-modal collaboration. These methods improve communication efficiency and can provide more abstract exchange formats than dense feature sharing. However, most still assume that sender and receiver were trained to participate in the same communication scheme. INTACT's protocol is asymmetric by design: the ego controls the request, the collaborator provides local evidence, and the ego decides whether and where to inject the response, enabling insertion of frozen heterogeneous collaborators without retraining the communication scheme.

\subsection{Sparse and Query-Based Interaction}
Sparse communication and query-based interaction have been widely explored in collaborative and transformer-based perception. Where2comm~\cite{hu2022where2comm} selects spatial regions for communication; deformable attention~\cite{zhu2020deformable} uses learned offsets to sample relevant features. These methods establish that perception can operate through selective evidence access. INTACT applies this principle at the cross-agent interface: typed evidence queries specify the location and type of evidence requested from a heterogeneous collaborator, responses are retrieved locally rather than densely translated, and routing serves as an efficiency-preserving selection mechanism. The key distinction is that sparsity in INTACT is not merely a way to reduce bandwidth within a compatible-feature framework; it is what makes local evidence exchange a viable alternative to global feature compatibility in the first place.
\section{Method}

\subsection{Overall Framework}
\begin{figure*}[t]
\centering
\includegraphics[width=0.98\textwidth]{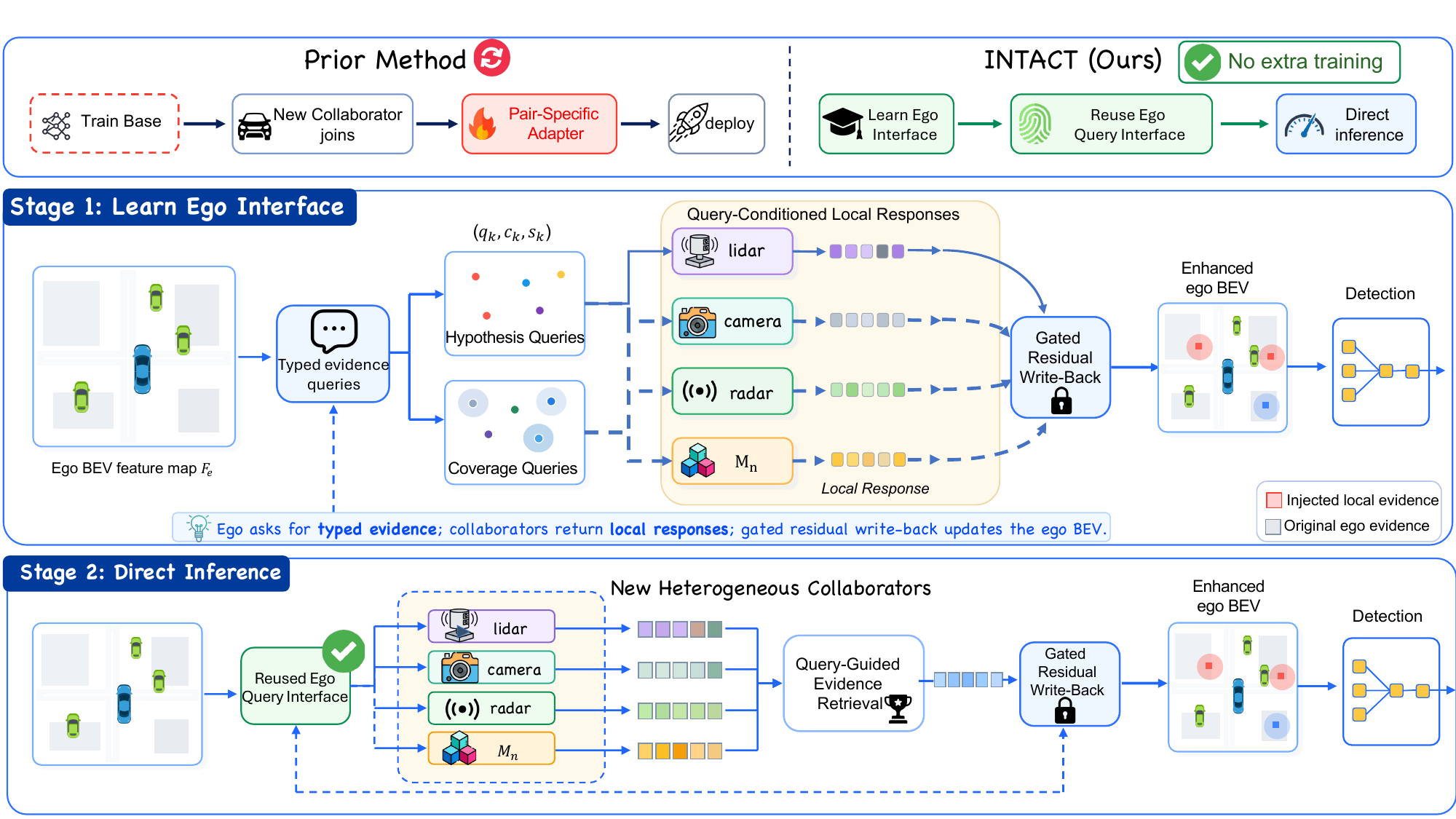}
\caption{Overview of INTACT. Stage 1 learns the ego-issued query interface once, including typed evidence query generation, query-conditioned local responses, Query-Guided Evidence Retrieval, and gated residual write-back. Stage 2 reuses the learned ego interface for direct inference with newly joined heterogeneous collaborators without extra pairwise heterogeneous adaptation.}
\label{fig:overview_intact}
\end{figure*}

As illustrated in Fig.~\ref{fig:overview_intact}, INTACT restructures heterogeneous collaborative perception from full-map feature translation into an ego-issued query interface. Prior methods typically require collaborators to first generate ego-interpretable representations through adapters, feature translators, or shared protocol spaces before participating in fusion. INTACT does not require collaborators to output complete ego-compatible feature maps. Instead, the ego issues typed evidence queries from its BEV representation. Each query specifies the location and type of evidence being requested, and collaborators return query-conditioned local responses only at the queried positions. The ego then retrieves and selects relevant local evidence through Query-Guided Evidence Retrieval and injects it into its own representation through gated residual write-back.

This process can be understood as a query, response, and gated residual write-back collaboration loop. The ego first issues local evidence requests, collaborators return local responses around those requests, and the ego injects useful evidence in a controlled manner. Unlike translation-first pipelines, the collaboration interface in INTACT is not defined by whether an entire collaborator feature map can be interpreted by the ego, but by what evidence the ego currently needs and where.

Given the ego feature map $F_e$, collaborator feature maps $\{F_j\}_{j=1}^{N-1}$, and their geometric transformations $\{T_{e\leftarrow j}\}_{j=1}^{N-1}$ to the ego coordinate system, INTACT first generates a set of typed evidence queries:
\begin{align}
    Q = \{(q_k, c_k, s_k)\}_{k=1}^{K},
\end{align}
where $q_k$ denotes query content, $c_k$ denotes the query location, and $s_k$ denotes query reliability, $N$ denotes the number of agents including the ego vehicle, and $K$ denotes the number of typed evidence queries issued by the ego. The index $j$ enumerates collaborators, while $k$ indexes ego-issued queries. The query set $Q$ contains two complementary request types: hypothesis queries request verification evidence around potential objects the ego already attends to, while coverage queries request supplementary evidence in occluded, distant, or low-response regions.

After query generation, the system retrieves local responses from collaborator features conditioned on each query, and selects the more relevant external evidence. Finally, the selected responses are reorganized as complementary residuals in the ego coordinate system and update the ego representation through gated residual write-back, yielding the enhanced feature $F_e^+$. The overall process is:
\begin{align}
    Q = G_q(F_e), \quad
    \{\tilde{v}_k\}_{k=1}^{K} = H\!\left(\{F_j, T_{e\leftarrow j}\}_{j=1}^{N-1}, Q\right), \quad
    F_e^+ = W\!\left(F_e, \{\tilde{v}_k\}_{k=1}^{K}, Q\right),
\end{align}
where $G_q$ denotes the typed evidence query generator, $H$ denotes the Query-Guided Evidence Retrieval module, and $W$ denotes the gated residual write-back module. Section~3.2 describes how the ego generates hypothesis queries and coverage queries. Section~3.3 describes how collaborators return query-conditioned local responses and how the ego retrieves and selects local evidence from candidate responses. Section~3.4 describes how selected external evidence is stably injected into the ego representation through gated residuals.

\subsection{Typed Evidence Query Generator}

INTACT does not start cross-vehicle interaction by directly exchanging dense features. The ego first generates typed evidence queries to express what local evidence it still needs from external collaborators given its current perceptual state. These queries are not direct copies of ego features; they are controlled evidence requests that specify where evidence is requested, what type of evidence is requested, and how reliable the request is in the current scene.

Specifically, given the ego feature map $F_e$, the model first produces intermediate features $Z_e = \psi(F_e)$ and predicts an object response map:
\begin{align}
    M = \sigma\!\left(h(Z_e)\right),
\end{align}
where $\psi$ is a lightweight ego-side feature projection, $h$ is an object-response head, and $\sigma$ denotes the sigmoid function.

This response map is not used for final detection. It serves as a spatial cue for query generation, indicating which locations are most likely to benefit from external evidence. $G_q$ selects high-response locations from this map as the positions of hypothesis queries, which request collaborators to verify potential objects the ego already attends to. However, if queries concentrate only in high-response regions, the interaction becomes overly biased toward locations where the ego is already relatively confident, making it difficult to actively cover occluded, distant, and low-response areas. To address this, INTACT further places a sparse set of fixed grid points on the normalized BEV plane as the positions of coverage queries to probe evidence-deficient regions. Hypothesis queries and coverage queries together form the final query location set, denoted $\{c_k\}_{k=1}^{K}$.
\begin{figure*}[t]
\centering
\includegraphics[width=0.98\textwidth]{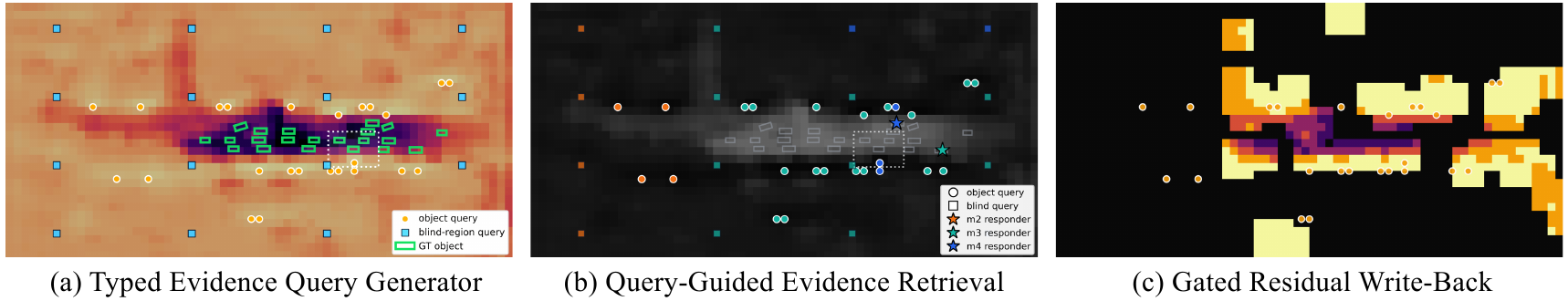}
\caption{Mechanism visualization of INTACT. (a) Typed Evidence Query Generator with a query-density heatmap, hypothesis queries, coverage queries, and ground-truth boxes. (b) Query-Guided Evidence Retrieval with query-conditioned local responses and color-coded collaborator assignments. (c) Gated Residual Write-Back showing the gated spatial residual injected into the ego feature map.}
\label{fig:intact_a5_visualization}
\end{figure*}

After determining query locations, $G_q$ constructs a typed evidence query for each location $c_k$. We define:
\begin{align}
    q_k = \mathrm{LN}\!\left(t_{\mathrm{task}} + t_{\mathrm{type}(k)} + \phi_{\mathrm{type}(k)}(p_k)\right) + P(c_k),
\end{align}
where $t_{\mathrm{task}}$ is a task token shared by all queries, $t_{\mathrm{type}(k)}$ distinguishes hypothesis queries from coverage queries, and $P(c_k)$ provides explicit positional encoding. $p_k$ is a lightweight local prior extracted by the query generator from the ego representation at location $c_k$, and $\phi_{\mathrm{type}(k)}(\cdot)$ maps this prior into the query space, modulating its contribution according to query type. This local prior supplies only the scene cues needed for query construction, such as object response or coverage deficiency signals, rather than directly copying the ego's full local features into the query. Consequently, $q_k$ expresses not what the ego already possesses at that location, but what local evidence the ego expects collaborators to supplement there under that query type.

In addition to query content and location, $G_q$ outputs a query score $s_k$ for each query, reflecting its reliability in the current scene. For hypothesis queries, this score is derived from the response magnitude at the corresponding location; for coverage queries, a fixed moderate confidence is assigned. This design lets subsequent modules know not only where to retrieve evidence and what type to retrieve, but also how strongly the returned evidence should participate in gated residual write-back. Finally, $G_q$ outputs the query set $(q_k, c_k, s_k)$, which is fed into the Query-Guided Evidence Retrieval module.

\subsection{Query-Guided Evidence Retrieval}

Once typed evidence queries are defined, the role of each collaborator is no longer to translate its entire feature map into a unified representation that the ego can directly fuse. Instead, the ego-issued query interface has already specified what the ego currently needs and where; collaborators need only return query-conditioned local responses at positions relevant to those requests.

Specifically, for each collaborator $j$, the system first geometrically transforms its feature map into the ego coordinate system:
\begin{align}
    \bar{F}_j = \mathrm{Warp}(F_j, T_{e\leftarrow j}).
\end{align}
It then extracts query-conditioned local responses at query locations $c_k$ via grid sampling:
\begin{align}
    v_{k,j} = \mathrm{Sample}(\bar{F}_j, c_k).
\end{align}
Here, $v_{k,j}$ is the local evidence returned by collaborator $j$ for query $k$. Importantly, this response comes only from the vicinity of the query anchor, not from a complete collaborator feature map. The module therefore does not require collaborators to produce full-map ego-compatible representations; it only requires that they can return comparable local responses around the ego's evidence requests.

After collecting candidate responses, the system does not directly average the local evidence from different collaborators. For the same query, different collaborators may return responses of varying quality, and relevance depends on local observations, modality capability, and the current query type. Averaging all responses indiscriminately would mix useful evidence with irrelevant or noisy contributions; heterogeneity would manifest as signal dilution. INTACT therefore adopts a per-query evidence selection mechanism. For each query, the system compares how well each collaborator's local response matches the current query and selects the more relevant evidence source.

To this end, queries and local responses are projected into a lightweight retrieval scoring space and matched via normalized similarity:
\begin{align}
    \hat{q}_k = \mathrm{Norm}(\eta_q(q_k)), \quad
    \hat{v}_{k,j} = \mathrm{Norm}(\eta_v(v_{k,j})), \quad
    r_{k,j} = \langle \hat{q}_k, \hat{v}_{k,j} \rangle.
\end{align}
Here, $\eta_q(\cdot)$ and $\eta_v(\cdot)$ are lightweight projections used solely for local evidence scoring. They do not translate a collaborator's entire feature map into an ego representation; they only establish the comparison space needed to match queries and local responses. $r_{k,j}$ indicates whether, under the local evidence request defined by query $k$, the response returned by collaborator $j$ is more suitable as the answer for that request. The system thus focuses not on how to translate entire feature maps, but on which collaborator's local evidence should be used for the current ego query.

During training, the system uses soft routing to distribute weights across collaborators, keeping the selection process differentiable:
\begin{align}
    a_{k,j} = \mathrm{softmax}_j(r_{k,j}/\tau_n), \quad
    \tilde{v}_k = \sum\nolimits_j a_{k,j}v_{k,j}.
\end{align}
Here, the softmax is computed over collaborators for each query $k$, $a_{k,j}$ is the routing weight for collaborator $j$, and $\tau_n$ controls the routing sharpness.

During inference, the system adopts a sparser winner-take-all strategy and retains only the most relevant local response:
\begin{align}
    j_k^* = \arg\max_j r_{k,j}, \quad
    \tilde{v}_k = v_{k,j_k^*}.
\end{align}
The former is a continuous relaxation of discrete collaborator selection; the latter better matches the deployment goal: each query retrieves evidence only from the most relevant source. After this stage, the originally dense collaborator feature maps have been compressed into a set of per-query local evidence vectors $\{\tilde{v}_k\}_{k=1}^{K}$. The remaining question is how to stably write these sparse evidence vectors back into the ego's main representation.

Regarding communication, the sparsity of INTACT comes from changing the unit of interaction: the system exchanges query-conditioned local responses rather than dense feature maps. In simulation, candidate local responses for each query are scored to evaluate how effectively the ego selects relevant evidence under query conditioning. In practical deployment, the same interface can be realized with a two-phase protocol: collaborators first return lightweight visibility or confidence cues, and the ego then requests local responses only from the selected sources. In either implementation, INTACT avoids full-map feature translation and dense feature transmission, rather than relying on indiscriminate fusion of all collaborator feature maps.

\subsection{Gated Residual Write-Back}
\begin{table*}[t]
\centering
\small
\setlength{\tabcolsep}{6pt}
\renewcommand{\arraystretch}{1.15}
\caption{Comparison on OPV2V-H and DAIR-V2X benchmarks. \#P(M) denotes heterogeneous enabling parameters per pair; Comm. Vol. is reported in $\log_2$ scale. Baseline values are from published results~\cite{mpda,heal,cofilling,stamp,gencomm} or local reproduction. The $\mathbf{L}_P^{64}$ ego-only detector, corresponding to $m_1$, serves as a strong LiDAR reference.}
\label{tab:main_results}
\resizebox{0.95\textwidth}{!}{%
\begin{tabular}{l|cc|cc|cc|cc|c|c}
\toprule
\multirow{3}{*}{Method}
& \multicolumn{6}{c|}{\cellcolor{gray!20}\textbf{OPV2V-H}}
& \multicolumn{2}{c|}{\cellcolor{gray!20}\textbf{DAIR-V2X}}
& \multirow{3}{*}{\makecell[c]{Comm.\\Vol.\\$(\log_2)\downarrow$}}
& \multirow{3}{*}{\makecell[c]{\#P(M)$\downarrow$}} \\
\cmidrule(lr){2-7} \cmidrule(lr){8-9}
&
\multicolumn{2}{c|}{$\mathbf{L}_{P}^{64}\text{-}\mathbf{C}_{E}$}
& \multicolumn{2}{c|}{$\mathbf{L}_{P}^{64}\text{-}\mathbf{L}_{S}^{32}$}
& \multicolumn{2}{c|}{$\mathbf{L}_{P}^{64}\text{-}\mathbf{C}_{R}$}
& \multicolumn{2}{c|}{$\mathbf{L}_{P}^{64}\text{-}\mathbf{L}_{S}^{40}$}
& & \\
& AP50 & AP70
& AP50 & AP70
& AP50 & AP70
& AP30 & AP50
& & \\
\midrule
MPDA~\cite{mpda}
& 0.8794 & 0.7520
& 0.8779 & 0.7459
& 0.8776 & 0.7464
& 0.4246 & 0.3641
& 22.0
& 1.91 \\
BackAlign~\cite{heal}
& 0.8807 & 0.7588
& \underline{0.9115} & \underline{0.7720}
& 0.8808 & 0.7579
& 0.4562 & 0.3727
& 22.0
& 21.84 \\
CodeFilling~\cite{cofilling}
& 0.8810 & 0.7565
& 0.8805 & 0.7580
& 0.8809 & 0.7583
& 0.3848 & 0.3189
& \textbf{14.7}
& \textbf{0.27} \\
STAMP~\cite{stamp}
& 0.8756 & \underline{0.7607}
& 0.8679 & 0.7576
& 0.8752 & \underline{0.7614}
& 0.4468 & \underline{0.3913}
& 22.0
& 1.09 \\
GenComm~\cite{gencomm}
& \underline{0.8953} & 0.7511
& \textbf{0.9241} & 0.7662
& \underline{0.8930} & 0.7508
& \underline{0.4593} & 0.3786
& \underline{16.1}
& 0.81 \\
\textbf{INTACT (Ours)}
& \textbf{0.9011} & \textbf{0.8009}
& 0.9015 & \textbf{0.8009}
& \textbf{0.9014} & \textbf{0.8008}
& \textbf{0.5049} & \textbf{0.4382}
& 18.0
& \underline{0.52} \\
\bottomrule
\end{tabular}%
}
\end{table*}

After the previous stage, the system possesses a set of query-indexed local external evidence vectors, rather than a shared representation aligned one-to-one with the ego feature map. INTACT therefore does not perform full-map fusion. Instead, it reorganizes the per-query evidence into complementary residuals in the ego coordinate system and injects them locally into the main representation through gated residual write-back.

First, the system uses the query score to modulate the strength of each selected response:
\begin{align}
    u_k = s_k\tilde{v}_k.
\end{align}
Here, $u_k$ denotes the reliability-weighted local evidence for query $k$, where $s_k$ is the query reliability predicted by the query generator and $\tilde{v}_k$ is the selected response from Query-Guided Evidence Retrieval.

This step does not alter the content of the returned evidence; it only controls how strongly that evidence should affect the subsequent gated residual write-back stage, preventing low-utility or low-confidence queries from causing overly strong updates.

Next, the system scatters these weighted responses back onto the ego BEV grid according to their query coordinates. If multiple queries fall on the same location, their responses are averaged, yielding a sparse residual field. Since this residual field remains discrete and sparse, INTACT further applies local convolutional encoding, normalization, and magnitude clipping to convert it into a writable dense complementary residual $\Delta$. This process does not aim to reconstruct a new collaborator feature map; it transforms sparse local answers into incremental updates in the ego coordinate system.

Finally, the system jointly uses the original ego feature $F_e$ and the complementary residual $\Delta$ to predict a spatial residual gate, and injects the residual as:
\begin{align}
    g = \mathrm{clip}_{[\rho,1]}\!\left(\sigma(\Gamma([F_e, \Delta]))\right), \quad
    F_e^+ = F_e + g \odot \Delta.
\end{align}
Here, $[F_e,\Delta]$ denotes channel-wise concatenation, $\Gamma(\cdot)$ is a lightweight gate prediction network, $\sigma(\cdot)$ is the sigmoid function, and $\mathrm{clip}_{[\rho,1]}(\cdot)$ constrains the gate value to the interval $[\rho,1]$. The operator $\odot$ denotes element-wise multiplication. The gate $g$ is a location-dependent residual injection strength, which decides how much complementary evidence should be written into each ego BEV location. The lower bound $\rho$ prevents the gate from collapsing too quickly to zero during early training, while the upper bound keeps the residual update bounded.

The key role of gated residual write-back is to separate the retrieval of external evidence from its safe injection into the ego representation. The preceding typed evidence queries define the evidence requests; Query-Guided Evidence Retrieval determines which local evidence to retrieve and select from collaborators; and gated residual write-back determines how this evidence enters the ego representation in a local and stable manner. Through this design, INTACT organizes heterogeneous collaboration as a complete ego-issued query interface: the ego issues typed evidence requests, collaborators return query-conditioned local responses, and the ego retrieves, selects, and injects useful evidence under controlled gating.
\begin{table*}[t]
\centering
\small
\setlength{\tabcolsep}{6pt}
\renewcommand{\arraystretch}{1.15}
\caption{Deployment-oriented results on V2X-Real and OPV2V-H. For V2X-Real, AP is reported on the vehicle class following the GenComm evaluation convention; $\mathbf{L}_H^{128}$ corresponds to the ego model $m_1$, and $\mathbf{L}_L^{128}$, $\mathbf{L}_M^{128}$, and $\mathbf{L}_T^{128}$ are sequentially inserted collaborators. The OPV2V-H block reports sequential multi-collaborator results, where $\mathbf{L}_P^{64}$ corresponds to $m_1$, together with the total deployed parameter count of the final multi-agent model.}
\label{tab:deployment_results}
\resizebox{0.95\textwidth}{!}{%
\begin{tabular}{l|cc|cc|cc|c}
\toprule
\rowcolor{gray!20}
\multicolumn{8}{c}{\textbf{V2X-Real Sequential Multi-Collaborator Setting}} \\
\midrule
\multirow{2}{*}{Method}
& \multicolumn{2}{c|}{$\mathbf{L}_H^{128}\text{+}\mathbf{L}_L^{128}$}
& \multicolumn{2}{c|}{$\mathbf{L}_H^{128}\text{+}\mathbf{L}_L^{128}\text{+}\mathbf{L}_M^{128}$}
& \multicolumn{2}{c|}{$\mathbf{L}_H^{128}\text{+}\mathbf{L}_L^{128}\text{+}\mathbf{L}_M^{128}\text{+}\mathbf{L}_T^{128}$}
& \multirow{2}{*}{\makecell[c]{\#P(M,total)$\downarrow$}} \\
& AP30 $\uparrow$ & AP50 $\uparrow$
& AP30 $\uparrow$ & AP50 $\uparrow$
& AP30 $\uparrow$ & AP50 $\uparrow$
& \\
\midrule
MPDA~\cite{mpda}
& 0.6344 & 0.5725
& 0.6323 & 0.5751
& 0.6211 & 0.5672
& 40.93 \\
BackAlign~\cite{heal}
& 0.6313 & 0.5822
& 0.6386 & 0.5878
& 0.6352 & 0.5896
& 39.02 \\
CodeFilling~\cite{cofilling}
& 0.6273 & 0.5826
& 0.6284 & 0.5799
& 0.6081 & 0.5571
& 29.35 \\
STAMP~\cite{stamp}
& 0.6314 & 0.5881
& 0.6335 & 0.5893
& 0.6289 & 0.5882
& 49.53 \\
GenComm~\cite{gencomm}
& \underline{0.6848} & \underline{0.6175}
& \underline{0.6961} & \underline{0.6299}
& \underline{0.7144} & \underline{0.6362}
& 46.76 \\
\textbf{INTACT (Ours)}
& \textbf{0.7083} & \textbf{0.6783}
& \textbf{0.7500} & \textbf{0.7116}
& \textbf{0.7588} & \textbf{0.7208}
& \textbf{25.40} \\
\midrule[0.8pt]
\rowcolor{gray!20}
\multicolumn{8}{c}{\textbf{OPV2V-H Sequential Multi-Collaborator Setting}} \\
\midrule
\multirow{2}{*}{Method}
& \multicolumn{2}{c|}{$\mathbf{L}_P^{64}\text{+}\mathbf{C}_E$}
& \multicolumn{2}{c|}{$\mathbf{L}_P^{64}\text{+}\mathbf{C}_E\text{+}\mathbf{L}_S^{32}$}
& \multicolumn{2}{c|}{$\mathbf{L}_P^{64}\text{+}\mathbf{C}_E\text{+}\mathbf{L}_S^{32}\text{+}\mathbf{C}_R$}
& \multirow{2}{*}{\makecell[c]{\#P(M,total)$\downarrow$}} \\
& AP50 $\uparrow$ & AP70 $\uparrow$
& AP50 $\uparrow$ & AP70 $\uparrow$
& AP50 $\uparrow$ & AP70 $\uparrow$
& \\
\midrule
MPDA~\cite{mpda}
& 0.8817 & 0.7449
& 0.8818 & 0.7451
& 0.8816 & 0.7444
& 40.93 \\
BackAlign~\cite{heal}
& 0.8818 & 0.7448
& 0.7600 & 0.6325
& 0.7601 & 0.6332
& 39.02 \\
CodeFilling~\cite{cofilling}
& 0.8826 & 0.7453
& 0.8828 & 0.7454
& 0.8826 & 0.7454
& 29.35 \\
STAMP~\cite{stamp}
& 0.8827 & 0.7461
& 0.8826 & 0.7468
& 0.8823 & 0.7481
& 49.53 \\
GenComm~\cite{gencomm}
& \underline{0.8941} & \underline{0.7486}
& \underline{0.9020} & \underline{0.7527}
& \underline{0.9018} & \underline{0.7531}
& 46.76 \\
\textbf{INTACT (Ours)}
& \textbf{0.9070} & \textbf{0.8003}
& \textbf{0.9070} & \textbf{0.8001}
& \textbf{0.9071} & \textbf{0.8002}
& \textbf{25.40} \\
\bottomrule
\end{tabular}%
}
\end{table*}

\section{Experiments}

\subsection{Experimental Settings}
\textbf{Datasets.} We evaluate INTACT on OPV2V-H~\cite{heal}, DAIR-V2X~\cite{dair_v2x}, and V2X-Real~\cite{v2xreal}. OPV2V-H is the primary simulated heterogeneous benchmark and supports controlled pairwise, no-train, and multi-collaborator evaluation. DAIR-V2X and V2X-Real provide real-world validation under more limited but practically important sensing conditions. Following prior heterogeneous collaborative perception work~\cite{heal}, we use $m_i$ as shorthand for concrete heterogeneous agent configurations. On OPV2V-H, $m_1=\mathbf{L}_P^{64}$ denotes the LiDAR PointPillars ego model with 64-channel BEV features, $m_2=\mathbf{C}_E$ denotes the camera EfficientNet collaborator, $m_3=\mathbf{L}_S^{32}$ denotes the LiDAR SECOND collaborator with 32-channel BEV features, and $m_4=\mathbf{C}_R$ denotes the camera ResNet101 collaborator. Thus the OPV2V-H pairwise columns $\mathbf{L}_P^{64}\text{-}\mathbf{C}_E$, $\mathbf{L}_P^{64}\text{-}\mathbf{L}_S^{32}$, and $\mathbf{L}_P^{64}\text{-}\mathbf{C}_R$ correspond to $m_1{+}m_2$, $m_1{+}m_3$, and $m_1{+}m_4$, respectively. On DAIR-V2X~\cite{dair_v2x}, we report the $\mathbf{L}_P^{64}\text{-}\mathbf{L}_S^{40}$ vehicle-infrastructure setting. For V2X-Real, we follow GenComm~\cite{gencomm} and denote $\mathbf{L}_H^{128}$ as the ego model $m_1$, while $\mathbf{L}_L^{128}$, $\mathbf{L}_M^{128}$, and $\mathbf{L}_T^{128}$ are the sequentially inserted heterogeneous collaborators. 

\textbf{Compared methods.} We compare with representative heterogeneous collaboration baselines that cover the major design families in this area. MPDA and BackAlign/HEAL learn alignment or adaptation modules for heterogeneous features. STAMP and GT-Space introduce shared interaction protocols. CodeFilling studies reconstruction or codebook-based interaction, while GenComm uses generative communication to support heterogeneous collaboration. These baselines retain a translation-first or protocol-compatibility premise in which collaborator representations must be made compatible before fusion. INTACT instead trains a train-once ego-issued query interface and reuses it for heterogeneous pairwise and multi-agent inference with no extra pairwise heterogeneous adaptation.

\textbf{Implementation details.} Training-dependent baselines follow their released homogeneous-to-heterogeneous pipelines. INTACT trains the ego-issued query interface once on $m_1$ and then inserts heterogeneous collaborators by reusing the same checkpoint. Unless otherwise specified, the remaining evaluation protocol follows GenComm. All experiments use four NVIDIA RTX 4090 GPUs with a total batch size of 4. Additional implementation details and deployment-oriented results are provided in the appendix.

\begin{table*}[t]
\centering
\caption{OPV2V-H results under Pyramid Fusion~\cite{heal}. All methods use Pyramid as the fusion backbone. INTACT keeps the ego-issued query interface, Query-Guided Evidence Retrieval, and gated residual write-back while replacing the default fusion backbone with Pyramid.}
\label{tab:pyramid_fusion}
\resizebox{0.95\textwidth}{!}{%
\begin{tabular}{c|l|cc|cc|cc|cc}
\toprule
\multirow{3}{*}{Fusion Network} & \multirow{3}{*}{Method} & \multicolumn{8}{c}{\cellcolor{gray!20}\textbf{OPV2V-H}} \\
\cmidrule(lr){3-10}
& & \multicolumn{2}{c|}{$\mathbf{L}_{S}^{64}$} 
& \multicolumn{2}{c|}{$\mathbf{L}_{P}^{64}\text{-}\mathbf{C}_{E}$}
& \multicolumn{2}{c|}{$\mathbf{L}_{P}^{64}\text{-}\mathbf{L}_{S}^{32}$}
& \multicolumn{2}{c}{$\mathbf{L}_{P}^{64}\text{-}\mathbf{C}_{R}$} \\
& & AP50 & AP70
& AP50 & AP70
& AP50 & AP70
& AP50 & AP70 \\
\midrule
\multirow{5}{*}{Pyramid Fusion} 
& Late fusion       & 0.818 & 0.690 & 0.772 & 0.615 & 0.821 & 0.680 & 0.766 & 0.613 \\
& HEAL~\cite{heal}              & 0.889 & 0.801 & 0.827 & 0.724 & 0.887 & 0.801 & 0.823 & 0.728 \\
& STAMP~\cite{stamp}             & 0.886 & 0.801 & 0.833 & 0.734 & 0.876 & 0.806 & 0.827 & 0.738 \\
& GT-Space~\cite{gtspace}          & \underline{0.894} & \underline{0.803} & \underline{0.848} & \underline{0.766} & \underline{0.891} & \underline{0.810} & \underline{0.844} & \underline{0.762} \\
& \textbf{INTACT (Ours)} & \textbf{0.922} & \textbf{0.833} & \textbf{0.895} & \textbf{0.810} & \textbf{0.899} & \textbf{0.815} & \textbf{0.895} & \textbf{0.812} \\
\bottomrule
\end{tabular}%
}
\end{table*}

\subsection{Main Results}
\textbf{Pairwise heterogeneous benchmarks.} Table~\ref{tab:main_results} reports pairwise heterogeneous results on OPV2V-H and DAIR-V2X. Across the three OPV2V-H heterogeneous pairs, INTACT obtains stable AP70 values of 0.8008 to 0.8010. This consistency is important because the collaborators differ in modality and backbone, yet the same ego-issued query interface is reused rather than retrained for each pair. The $m_1$ ego-only result should be interpreted as a strong LiDAR reference, not as the main collaboration baseline that every heterogeneous pair must surpass. Under this reference, the small gap between the $m_1$ single-agent detector and INTACT indicates that inserting weaker or different collaborators does not introduce substantial negative transfer.

On DAIR-V2X, INTACT reaches AP30 0.5049 and AP50 0.4382 under the listed setting, outperforming the compared baselines in Table~\ref{tab:main_results}. This result suggests that typed evidence queries can remain useful beyond the simulated OPV2V-H setting, although real-world validation should still be expanded. INTACT also keeps the heterogeneous enabling overhead small: it adds 0.52M parameters per pair, compared with 21.84M for BackAlign and 0.81M for GenComm. Its communication volume is 18.0 $\log_2$, which corresponds to roughly 16$\times$ compression relative to dense feature transmission. Thus, the main table supports the intended trade-off: stable heterogeneous collaboration with substantially lower pair-specific parameter overhead than alignment-heavy baselines.

\textbf{V2X-Real and deployment evidence.} Table~\ref{tab:deployment_results} (top) reports V2X-Real vehicle-class AP following the GenComm evaluation convention. INTACT reaches AP50 values of 0.6783, 0.7116, and 0.7208 across the listed V2X-Real collaborator settings, outperforming the compared baselines while using the smallest deployed parameter count among the compared methods. These real-data results support the same deployment-oriented trade-off observed on OPV2V-H and DAIR-V2X: INTACT keeps a lightweight ego-issued query interface without adding pair-specific heterogeneous translators.

Table~\ref{tab:deployment_results} (bottom) further reports OPV2V-H sequential multi-collaborator deployment results. INTACT maintains strong AP50/AP70 as additional heterogeneous collaborators are inserted, while using the smallest total deployed parameter count among the compared methods. Table~\ref{tab:pyramid_fusion} evaluates INTACT with Pyramid as the fusion backbone and shows that the typed evidence query interface remains runnable under a different fusion backbone.

\subsection{Qualitative Detection Comparison}
\begin{figure*}[t]
\centering
\includegraphics[width=0.98\textwidth]{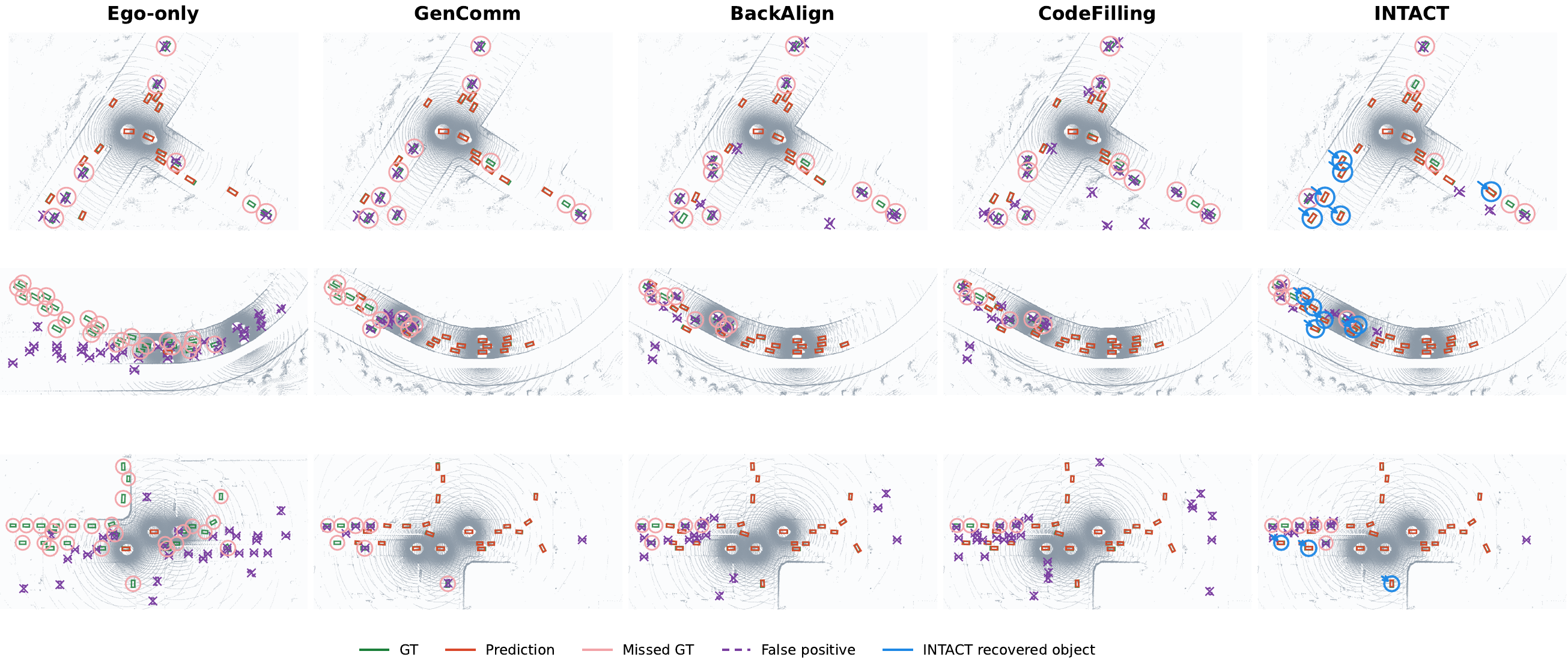}
\caption{
Qualitative detection comparison on the challenging OPV2V-H $m_1 \rightarrow m_4$ heterogeneous setting.
We compare Ego-only, GenComm, BackAlign, CodeFilling, and INTACT on three representative scenes with identical BEV crops and visualization settings.
Green boxes denote ground-truth objects and red boxes denote predictions.
Pink circles highlight missed ground-truth objects, purple dashed marks indicate false positives, and blue circles mark objects recovered by INTACT.
The examples illustrate that INTACT retrieves useful complementary evidence in difficult heterogeneous cases, leading to more complete detections and fewer visually apparent errors.
}
\label{fig:qualitative_detection}
\end{figure*}
\begin{figure*}[t]
\centering
\includegraphics[width=\textwidth]{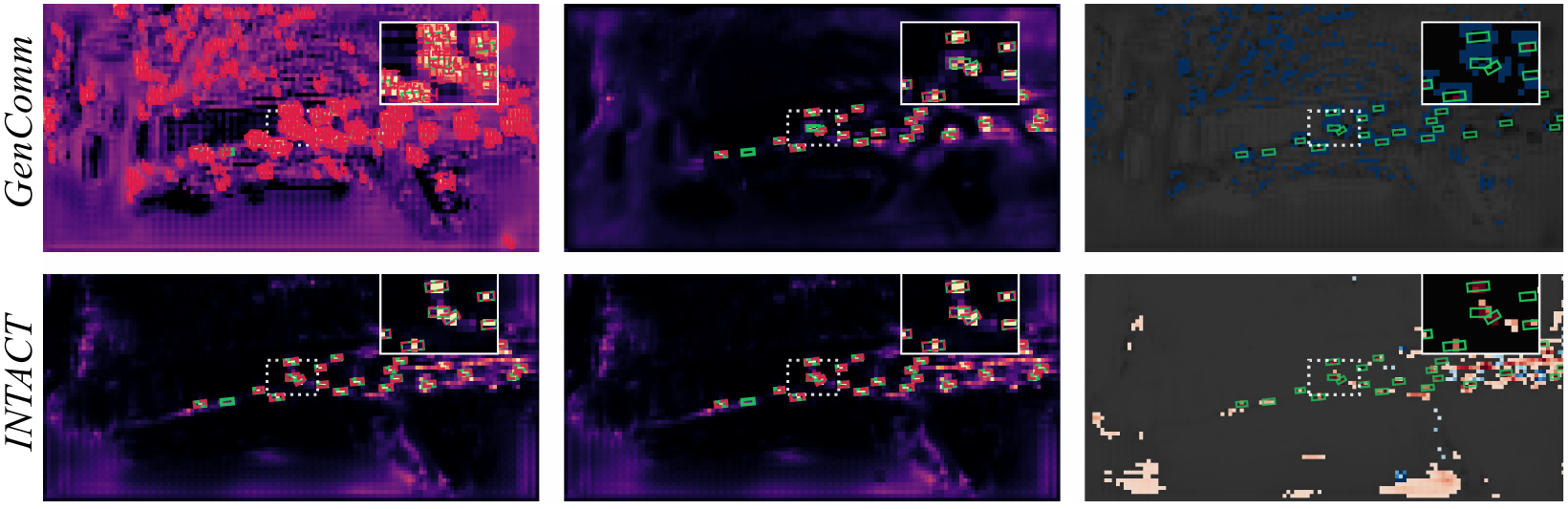}
\caption{Qualitative no-train comparison between GenComm and INTACT on the same OPV2V-H $m_1 \rightarrow m_2$ scene. Each row shows the ego response, the direct no-train pairwise response, and the signed gain. INTACT produces more target-concentrated positive updates, which is consistent with its train-once ego-issued query interface.}
\label{fig:notrain_compare}
\end{figure*}

Fig.~\ref{fig:qualitative_detection} provides a BEV qualitative comparison on the challenging OPV2V-H $m_1 \rightarrow m_4$ ($\mathbf{L}_P^{64}\rightarrow\mathbf{C}_R$) setting. All methods are visualized on the same scenes with identical crops and display settings, so differences in missed objects and false positives can be compared directly. The first two rows show high-gap cases where INTACT recovers objects that multiple baselines miss, while the third row is a more neutral representative case. This mix avoids presenting only success cases while still exposing the failure modes that motivate typed evidence retrieval.

The qualitative behavior is consistent with the mechanism of INTACT. GenComm, BackAlign, and CodeFilling often produce either missed ground-truth objects or additional false positives in crowded, occluded, or long-range regions. INTACT's blue highlighted detections indicate cases where query-conditioned local responses provide complementary evidence at locations that the ego or other baselines handle less reliably. These examples do not by themselves replace quantitative evaluation, but they illustrate how the ego-issued query interface can turn sparse collaborator evidence into localized improvements through Query-Guided Evidence Retrieval and gated residual write-back.

\subsection{Ablation Studies}

Table~\ref{tab:ablation_core} reports ablation studies on OPV2V-H across the three pairwise heterogeneous settings. We evaluate the contribution of three components in the interaction chain: typed evidence queries, Query-Guided Evidence Retrieval, and gated residual write-back.

Removing typed evidence queries leads to the largest degradation. In the $m_1{+}m_2$ setting, AP70 drops from 0.8009 to 0.6374, showing that untyped interaction is insufficient when the ego must decide where and what evidence to request from heterogeneous collaborators. The degradation is smaller but still visible on $m_1{+}m_3$ and $m_1{+}m_4$, where AP70 decreases from 0.8009 to 0.7653 and from 0.8008 to 0.7726, respectively. These results indicate that typed evidence queries are the most important component on average, especially under the $m_1{+}m_2$ setting, while gated residual write-back provides a more uniform stabilization effect across all pairs.

Gated residual write-back is also important for stable evidence injection. Replacing the learned gate with an ungated residual update reduces AP70 from about 0.8008--0.8009 to 0.7654--0.7656 across all three pairs. This consistent drop suggests that external evidence should not be directly added to the ego representation without spatially adaptive control. The gate helps determine where the retrieved evidence should affect the ego BEV feature map and prevents noisy or weak responses from dominating the update.

Removing Query-Guided Evidence Retrieval causes a smaller but consistent drop, with AP70 decreasing to 0.7933--0.7934 across the three pairwise settings. This shows that the model can still retain much of its performance when typed queries and gated write-back remain active, but per-query evidence retrieval further improves the use of collaborator responses. Overall, the ablation results support the design of INTACT: typed evidence queries provide the main interface for heterogeneous evidence requests, Query-Guided Evidence Retrieval selects useful local responses, and gated residual write-back stabilizes their injection into the ego representation.

\subsection{More Analysis}

\begin{table*}[t]
\centering
\footnotesize
\setlength{\tabcolsep}{3.8pt}
\renewcommand{\arraystretch}{1.12}
\caption{Ablation studies on OPV2V-H. We report pairwise results between $m_1$ and the other three heterogeneous collaborators.}
\label{tab:ablation_core}
\resizebox{0.98\textwidth}{!}{%
\begin{tabular}{lccc cc c cc c cc}
\toprule
\multirow{2}{*}{\textbf{Variant}} &
\multicolumn{3}{c}{\textbf{Components}} &
\multicolumn{2}{c}{\textbf{$\mathbf{m1 + m2}$}} &&
\multicolumn{2}{c}{\textbf{$\mathbf{m1 + m3}$}} &&
\multicolumn{2}{c}{\textbf{$\mathbf{m1 + m4}$}} \\
\cmidrule(lr){2-4}
\cmidrule(lr){5-6}
\cmidrule(lr){8-9}
\cmidrule(lr){11-12}
& \makecell{Typed\\queries}
& \makecell{Q-cond.\\response}
& \makecell{Gated\\write-back}
& AP50 & AP70
&& AP50 & AP70
&& AP50 & AP70 \\
\midrule

Full INTACT
& \checkmark & \checkmark & \checkmark
& 0.9011 & 0.8009
&& 0.9015 & 0.8009
&& 0.9014 & 0.8008 \\

w/o query-guided evidence retrieval
& \checkmark &  & \checkmark
& 0.8920 & 0.7934
&& 0.8920 & 0.7933
&& 0.8920 & 0.7934 \\

w/o gated residual write-back
& \checkmark & \checkmark & 
& 0.8873 & 0.7656
&& 0.8876 & 0.7655
&& 0.8879 & 0.7654 \\

w/o typed evidence queries
&  & \checkmark & \checkmark
& 0.7792 & 0.6374
&& 0.8462 & 0.7653
&& 0.8774 & 0.7726 \\
\bottomrule
\end{tabular}%
}
\end{table*}

\begin{figure*}[t]
    \centering
    \begin{minipage}[t]{0.56\linewidth}
        \centering
        \includegraphics[width=\linewidth]{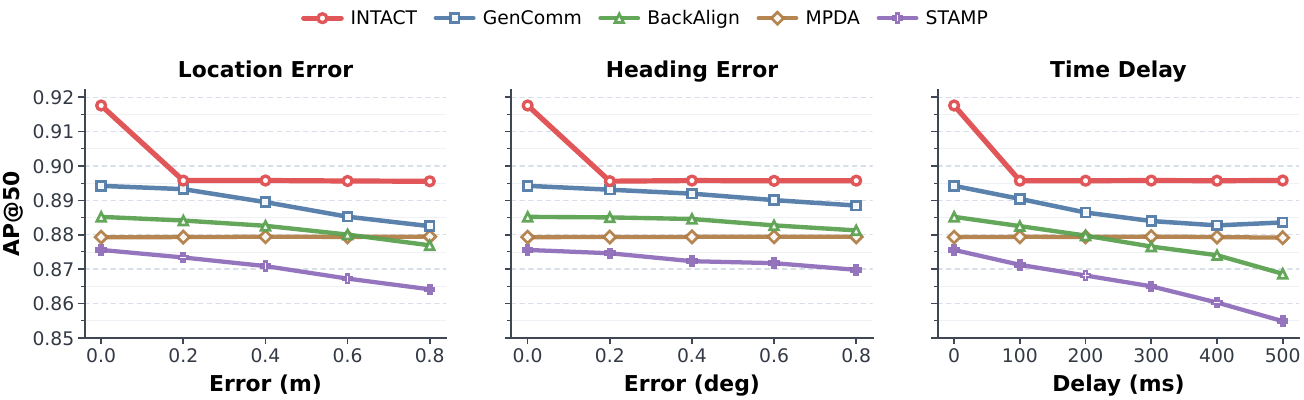}
    \end{minipage}
    \hfill
    \begin{minipage}[t]{0.43\linewidth}
        \centering
        \includegraphics[width=\linewidth]{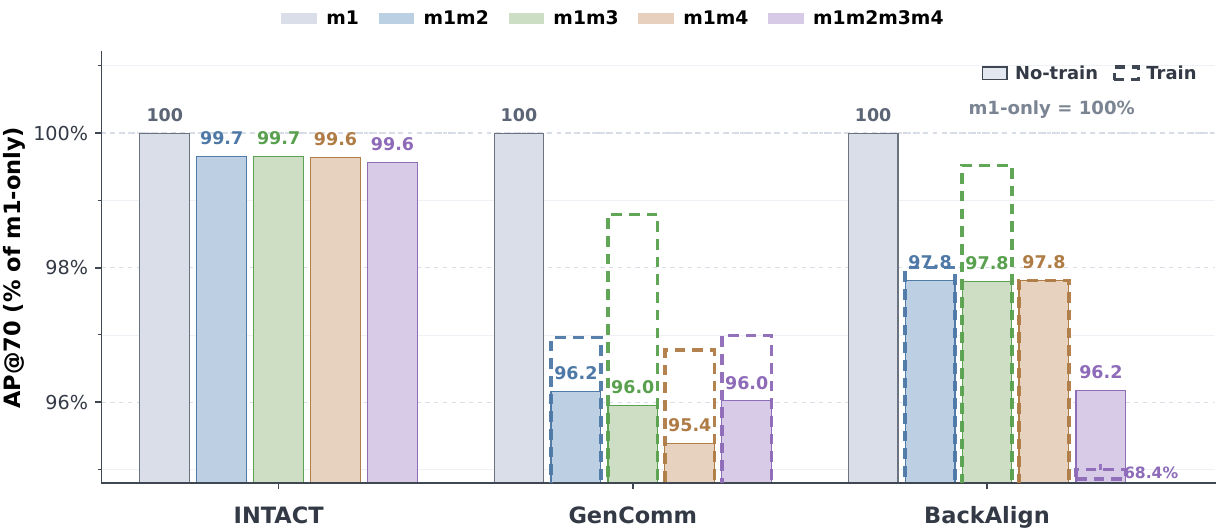}
    \end{minipage}
    \caption{Robustness and no-train retention analysis on OPV2V-H. Left: detection performance under location errors, heading errors, and time delays. Right: solid bars show AP70 retention under direct no-train collaborator switching, computed from Appendix Table~\ref{tab:absolute_notrain} as $\mathrm{AP70}(\mathrm{setting}) / \mathrm{AP70}(m_1)$. Dashed outlines indicate the corresponding trained reference when available. INTACT maintains higher no-train retention under collaborator switching, supporting the train-once ego-issued query interface.}
    \label{fig:more_analysis}
\end{figure*}
\begin{figure}[t]
    \centering
    \includegraphics[width=0.98\linewidth]{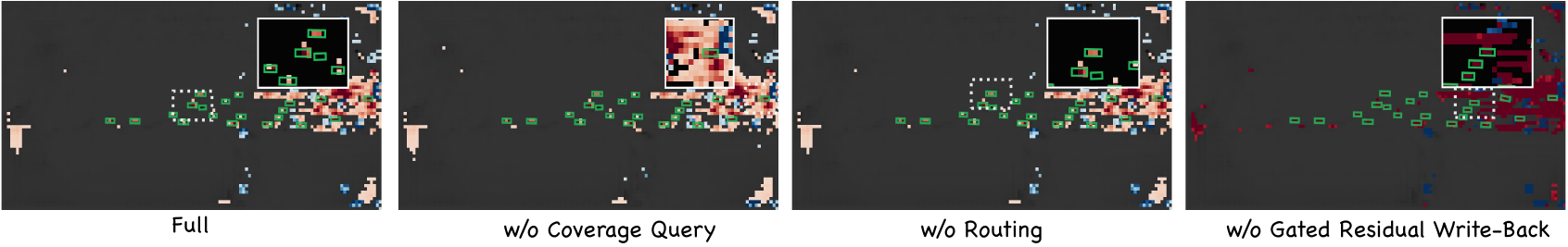}
    \caption{Qualitative sensitivity visualization on the hardest OPV2V-H $m_1 \rightarrow m_4$ pair, complementary to the retrained staircase in Table~\ref{tab:ablation_core}. Compared with the full model, suppressing components of the interaction chain at inference weakens the spatial concentration of useful responses, with the most visible disruption caused by removing the gated residual write-back. This panel illustrates inference-time spatial behavior and is not a one-to-one match to the retrained quantitative variants.}
    \label{fig:ablation_vis}
\end{figure}

\textbf{Robustness.} Fig.~\ref{fig:more_analysis} (left) evaluates robustness under pose errors and time delays. All methods degrade as the geometric or temporal mismatch increases, which is expected for cooperative perception. INTACT remains comparatively stable in the plotted curves, especially under larger perturbations where training-dependent baselines drop more sharply.

\textbf{No-train retention behavior.} Fig.~\ref{fig:more_analysis} (right) compares AP70 retention when pretrained models are directly switched to untrained heterogeneous collaborator settings, with each method's $m_1$ reference normalized to 100\%. INTACT stays close to its $m_1$ reference across all collaborator compositions, whereas GenComm and BackAlign suffer larger retention drops. In particular, BackAlign degrades sharply in the full multi-collaborator setting, retaining only 68.4\%. This contrast supports the claim that the train-once ego-issued query interface is more stable under collaborator switching than pair-specific heterogeneous adaptation.

\textbf{Qualitative ablation.} Fig.~\ref{fig:ablation_vis} visualizes inference-time mechanism suppression on the hardest OPV2V-H $m_1 \rightarrow m_4$ pair. The full model produces more spatially concentrated target-related responses, while suppressing the interaction chain disperses them, with the most disruptive spatial pattern when the gated residual write-back is removed. This qualitative panel should be read as a sensitivity visualization alongside the retrained staircase in Table~\ref{tab:ablation_core}; it illustrates spatial behavior rather than reproducing the retrained quantitative variants.

\section{Conclusion}

We presented INTACT, an ego-guided typed sparse evidence retrieval framework for heterogeneous collaborative perception. Instead of making collaborator features globally ego-compatible before fusion, INTACT defines collaboration through typed evidence queries issued by the ego vehicle. Collaborators return query-conditioned local responses at queried locations, and the ego injects selected evidence through gated residual write-back. This design shifts heterogeneous collaboration from full-map feature translation to local evidence retrieval under an ego-issued query interface. Experiments across simulated and real-world benchmarks show that INTACT achieves competitive detection accuracy with low communication and parameter overhead, while supporting a train-once deployment pattern without extra pairwise heterogeneous adaptation. The retrained staircase ablation further indicates that typed evidence queries are central to the framework, with gated residual write-back providing additional stabilization for injecting retrieved evidence. INTACT still assumes reasonably accurate geometric alignment and an ego-centered request-and-injection protocol. Future work should extend typed evidence retrieval to broader real-world settings, richer collaborator selection, stronger temporal reasoning, and more diverse ego and collaborator modalities.

{
\bibliographystyle{ieeenat_fullname}
\bibliography{main}
}

\appendix
\section{Technical appendices and supplementary material}

\subsection{Additional Implementation Details}
Unless otherwise stated, all training-dependent baselines follow the released homogeneous-to-heterogeneous training schedules of GenComm/HEAL. For INTACT on OPV2V-H, the stage-1 INTACT query-interface setting uses a 31-epoch schedule with Adam and an initial learning rate of $10^{-3}$. The default query configuration uses up to 512 queries with a hypothesis ratio of 0.75; coverage queries are sampled from a deficiency-based score map with an $8\times8$ anchor grid; routing uses soft assignment during training and hard selection at inference; and the gated residual write-back uses a gate floor of 0.08 with delta clipping at 5.0. Pairwise and multi-agent heterogeneous evaluations then reuse the learned $m_1$ checkpoint directly, without introducing an extra hetero retraining stage. Real-data experiments follow the same pipeline, with dataset-specific sensor/backbone definitions kept identical to the official benchmark settings

\subsection{Supplementary Deployment-Oriented Results}

\begin{table*}[h]
\centering
\small
\setlength{\tabcolsep}{5pt}
\renewcommand{\arraystretch}{1.12}
\caption{Absolute no-train deployment performance on OPV2V-H. Each method starts from its own pretrained $m_1$ ego and directly switches to the listed collaborators without any extra heterogeneous retraining. This table complements the no-train retention panel in Figure~\ref{fig:more_analysis} by reporting absolute AP rather than normalized retention.}
\label{tab:absolute_notrain}
\resizebox{0.98\textwidth}{!}{%
\begin{tabular}{l|cc|cc|cc|cc|cc}
\toprule
\multirow{2}{*}{Method}
& \multicolumn{2}{c|}{$m1$}
& \multicolumn{2}{c|}{$m1 \rightarrow m2$}
& \multicolumn{2}{c|}{$m1 \rightarrow m3$}
& \multicolumn{2}{c|}{$m1 \rightarrow m4$}
& \multicolumn{2}{c}{$m1 \rightarrow m1m2m3m4$} \\
\cmidrule(lr){2-3}\cmidrule(lr){4-5}\cmidrule(lr){6-7}\cmidrule(lr){8-9}\cmidrule(lr){10-11}
& AP50 & AP70 & AP50 & AP70 & AP50 & AP70 & AP50 & AP70 & AP50 & AP70 \\
\midrule
GenComm & \textbf{0.9300} & 0.7758 & 0.8853 & 0.7460 & 0.8828 & 0.7444 & 0.8737 & 0.7400 & 0.8899 & 0.7448 \\
BackAlign & \underline{0.9212} & 0.7749 & 0.8807 & \underline{0.7580} & 0.8806 & \underline{0.7578} & 0.8807 & \underline{0.7579} & 0.8827 & \underline{0.7453} \\
\rowcolor{gray!12}\textbf{INTACT (Ours)} & 0.9108 & \textbf{0.8037} & \textbf{0.9011} & \textbf{0.8009} & \textbf{0.9015} & \textbf{0.8009} & \textbf{0.9014} & \textbf{0.8008} & \textbf{0.9071} & \textbf{0.8002} \\
\bottomrule
\end{tabular}%
}
\end{table*}

\subsection{Deployment-Oriented Protocol Comparison}
\begin{table*}[h]
\centering
\small
\setlength{\tabcolsep}{5pt}
\renewcommand{\arraystretch}{1.15}
\caption{Workflow-level comparison of heterogeneous collaboration protocols when collaborator composition changes. The table summarizes each method's intended deployment workflow, rather than whether a manually merged checkpoint can be forced to run once.}
\label{tab:deployment_protocol}
\resizebox{0.98\textwidth}{!}{%
\begin{tabular}{l >{\raggedright\arraybackslash}p{3.0cm} >{\raggedright\arraybackslash}p{3.6cm} >{\centering\arraybackslash}p{1.6cm} >{\centering\arraybackslash}p{1.7cm} >{\raggedright\arraybackslash}p{2.4cm}}
\toprule
\multirow{2}{*}{Method} & \multicolumn{2}{c}{\textbf{Training Dependency}} & \multicolumn{2}{c}{\textbf{Deployment Behavior}} & \multirow{2}{*}{\textbf{Interface Paradigm}} \\
\cmidrule(lr){2-3} \cmidrule(lr){4-5}
 & \textbf{Before Hetero Deployment} & \textbf{When a New Collaborator Joins} & \textbf{Checkpoint} & \textbf{Switch Cost} &  \\
\midrule
\rowcolor{gray!12} \textbf{INTACT} & Reuse stage-1 base & None; reuse the same ego-issued query interface & Existing stage-1 base & \textbf{Direct} & Ego-typed evidence query \\
GenComm & Train method-specific homogeneous generators before hetero integration & Retrain deformable message extractor for the new pair/combo & Pair/combo checkpoint & Retrain & Shared generator \\
BackAlign & Reuse stage-1 base & Retrain the new-agent alignment branch / encoder side & Pair checkpoint & Retrain & Pairwise alignment \\
MPDA & Reuse stage-1 base & Retrain the hetero adaptation module for the new combo & Combo checkpoint & Retrain & Domain adaptation \\
CodeFilling & Reuse stage-1 base & Retrain the codebook / fusion module for the new combo & Combo checkpoint & Retrain & Code reconstruction \\
STAMP & Reuse stage-1 base & Retrain the protocol adapter and reverter & Pair checkpoint & Retrain & Shared protocol \\
\bottomrule
\end{tabular}%
}
\end{table*}

Table~\ref{tab:deployment_protocol} complements the no-train analyses in the main paper by making the workflow difference explicit. Existing baselines are still organized around pair- or combo-specific heterogeneous training steps, which means that when collaborator composition changes, the system must prepare a new checkpoint or retrain a dedicated interaction module. INTACT instead reuses the same ego-issued query interface and therefore supports direct collaborator insertion in its intended workflow. This workflow-level difference is the deployment counterpart to the main-paper no-train retention and absolute no-train utility results.

\end{document}